\documentclass[runningheads]{llncs}

\usepackage{eccv}

\usepackage{eccvabbrv}
\usepackage{multicol}
\usepackage{multirow}
\usepackage{graphicx}
\usepackage{booktabs}

\usepackage[accsupp]{axessibility}  

\usepackage[pagebackref,breaklinks,colorlinks,citecolor=eccvblue]{hyperref}

\usepackage{orcidlink}

\begin{document}

\title{Watt for What: Rethinking Deep Learning's Energy-Performance Relationship} 


\author{Shreyank N Gowda\inst{1} \and
Xinyue Hao\inst{2} \and
Gen Li\inst{2} \and
Shashank Narayana Gowda\inst{3} \and
Xiaobo Jin\inst{4} \and
Laura Sevilla-Lara\inst{2} }

\authorrunning{S. N. Gowda et al.}

\institute{University of Oxford, UK \\
\email{shreyank.narayanagowda@eng.ox.ac.uk}
\and
University of Edinburgh, UK \\
\and
UCLA, USA\\
\and
Xi'an Jiaotong-Liverpool University, China\\
}

\maketitle

\begin{abstract}
Deep learning models have revolutionized various fields, from image recognition to natural language processing, by achieving unprecedented levels of accuracy. However, their increasing energy consumption has raised concerns about their environmental impact, disadvantaging smaller entities in research and exacerbating global energy consumption. In this paper, we explore the trade-off between model accuracy and electricity consumption, proposing a metric that penalizes large consumption of electricity. We conduct a comprehensive study on the electricity consumption of various deep learning models across different GPUs, presenting a detailed analysis of their accuracy-efficiency trade-offs. We propose a metric that evaluates accuracy per unit of electricity consumed, demonstrating how smaller, more energy-efficient models can significantly expedite research while mitigating environmental concerns. Our results highlight the potential for a more sustainable approach to deep learning, emphasizing the importance of optimizing models for efficiency. This research also contributes to a more equitable research landscape, where smaller entities can compete effectively with larger counterparts. This advocates for the adoption of efficient deep learning practices to reduce electricity consumption, safeguarding the environment for future generations whilst also helping ensure a fairer competitive landscape. 
\end{abstract}

\section{Introduction}
\label{sec:intro}

Deep learning has emerged as a powerful technology, achieving remarkable breakthroughs across various domains. From image recognition and natural language processing to autonomous driving and healthcare diagnostics, deep learning models have redefined the boundaries of what machines can accomplish. The stunning advances in accuracy have transformed industries, offering new solutions to long-standing problems. Yet, advancements in deep learning carry a heavy toll, evident in their escalating energy consumption \cite{desislavov2021compute} and environmental impact \cite{selvan2022carbon}. As models expand in size and complexity, they require more computational resources, driving up operational costs and environmental concerns \cite{anthony2020carbontracker, strubell2019energy}.

In a landscape where cutting-edge deep learning research often necessitates access to colossal computational infrastructure, small companies and academic institutions find themselves at a disadvantage. Competing with tech giants and well-funded organizations on the basis of computational resources alone is an unattainable goal for many. This stark inequality not only hampers innovation but also perpetuates a power imbalance in the field, potentially biasing progress toward the economic interests of these dominant companies. 

\begin{figure}
    \centering
    \includegraphics[width=0.75\linewidth]{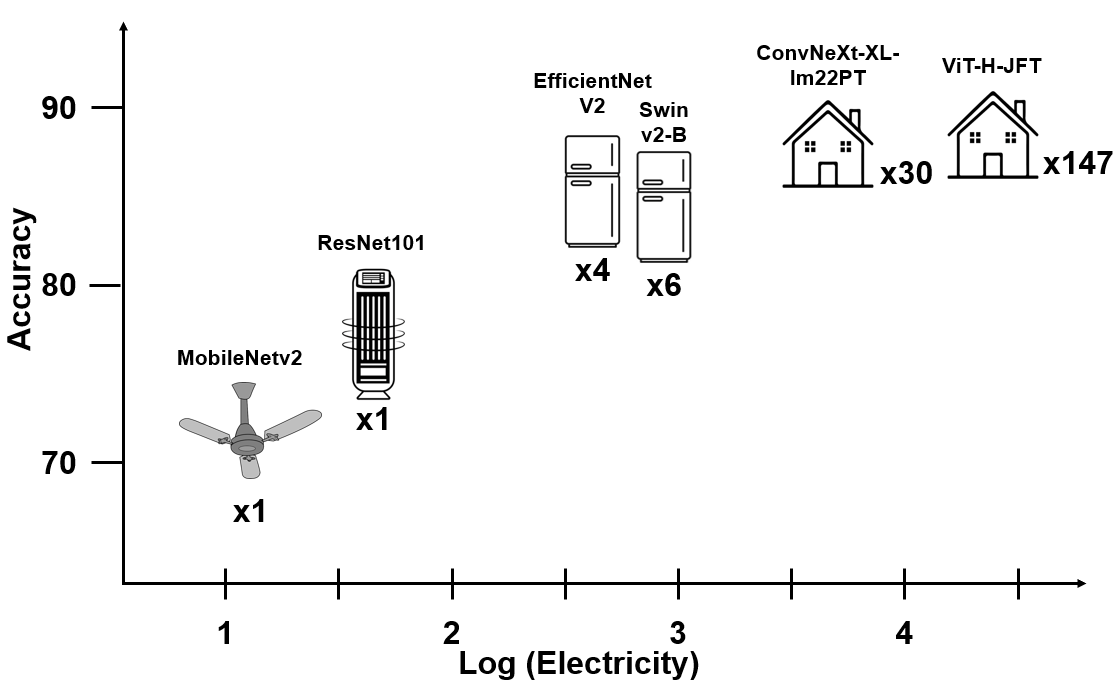}
    \caption{Bridging the Energy Divide: Deep Learning Models vs. Everyday Power Hogs. For easy comparison, we list the amount of electricity consumed per month by an appliance or by the average household in the UK.}
\label{fig:teaser}    
\end{figure}

The environmental crisis is worsening, with climate change accelerating, species disappearing rapidly, and pollution endangering human health globally. Deep learning and AI growth are aggravating these issues. Training complex AI models demands vast data and computing power, consuming massive electricity. For instance, a single large language model can emit as much carbon as a car over its lifetime \cite{bannour2021evaluating}. As these models expand, their energy consumption and carbon emissions soar. The OPT-175B Meta model alone consumes 356,000 kWh during training, while GPT-3-175B model uses a staggering 1,287,000 kWh \cite{khowaja2023chatgpt}. 

This unsustainable trend clashes with the urgent need to reduce emissions and shift to clean energy sources. While AI and deep learning offer substantial benefits, their development must prioritize energy efficiency to mitigate environmental impact. This paper takes a crucial step toward a more balanced and sustainable deep learning approach. We explore the trade-off between model accuracy and electricity consumption, introducing a metric to level the playing field in deep learning research. Our investigation spans from older models \cite{vgg16, resnet, colornet} to the latest transformers \cite{vit,deit,swin}, including self-supervised \cite{mae,clip} learning models. We extend our analysis beyond image classification to image segmentation and action recognition, considering model training and pre-training costs.

By assessing accuracy per unit of electricity, we level the playing field for smaller entities like universities and startups to compete with industry giants. We also endorse the adoption of energy-efficient model architectures, which not only cut electricity usage but also accelerate research, boosting overall efficiency. In Figure~\ref{fig:teaser}, we juxtapose deep learning model costs with real-life electricity estimates, comparing them to daily appliances and, for larger models, estimating them against household consumption \cite{avghousehold}. This paper offers a comprehensive analysis of deep learning model efficiency and its impact on research, industry, and the environment. It underscores the necessity for a fundamental shift in how we gauge and enhance deep learning models, emphasizing sustainability.

\section{Related Work}
\label{sec:related}

\subsection{Works about efficiency}

\begin{figure*}
    \centering
    \includegraphics[width=0.75\textwidth]{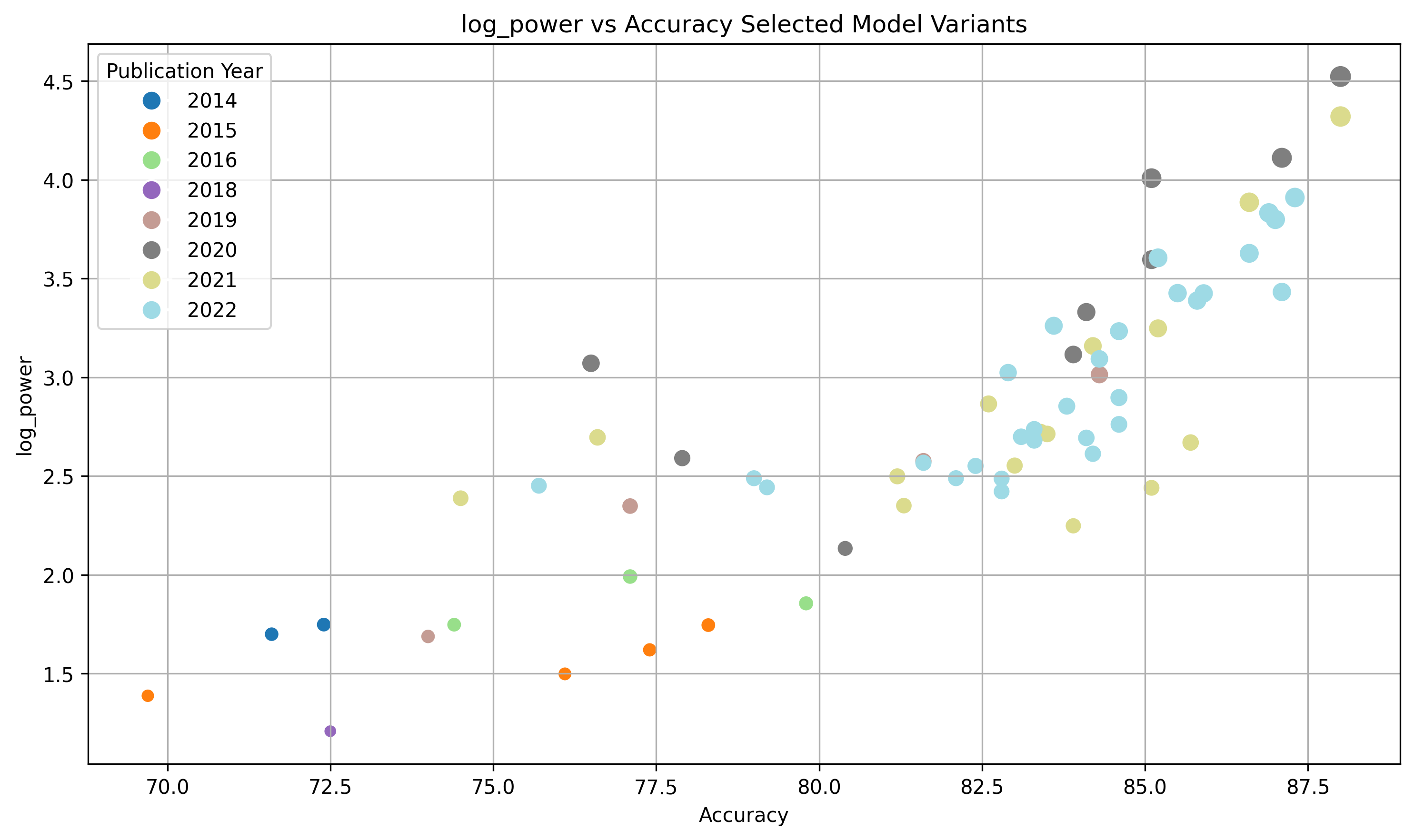}
    \caption{Comparing accuracy and electricity consumption over time. We see that whilst the accuracy of models are improving at a linear rate, the electricity consumed to train them is exponential. The electricity is measured in kWh.}
    \label{fig:con}
\end{figure*}

Enhancing neural network efficiency is a vital focus in deep learning research. As networks grow larger, their computational demands skyrocket. Figure \ref{fig:con} illustrates how small accuracy gains lead to exponential increases in electricity consumption. To address this, pruning \cite{pruning} removes redundant connections, while quantization \cite{quantization} reduces precision without sacrificing much accuracy and recent parameter-efficient finetuning strategies such as adapters~\cite{adapters} and LoRA have been proposed. Efficient architectures like Xception \cite{xception} and MobileNets \cite{mobilenet}, and techniques like knowledge distillation \cite{kd}, offer further efficiency gains.

For video understanding, where computations are intensive, methods like temporal shift modules \cite{lin2019tsm}, selective frame prediction \cite{scsampler,smart}, compressed temporal reasoning~\cite{kim2022capturing,dynamicimages} offer efficiency. However, defining efficiency is complex, considering factors like FLOPs, inference time, and memory usage \cite{efficiencymisnomer}. Improvements in one area may not translate to overall efficiency, necessitating careful consideration of specific goals and hardware constraints.

\subsection{Environmental Impact}

Research on the environmental impact of deep learning models is a burgeoning field. Ligozat et al. \cite{ligozat2022unraveling} review tools for assessing AI's environmental impacts, advocating for energy-efficient algorithms and hardware. However, this work lacks empirical data. Large deep learning models incur significant carbon emissions and energy usage during training \cite{strubell2019energy}. For instance, training a large transformer-based language model can emit 626,000 pounds of CO2, equivalent to 5 times the lifetime emissions of an average American car \cite{strubell2019energy}. Efforts to reduce the carbon footprint of deep learning include techniques like early stopping and model card documentation to detail compute infrastructure and emissions \cite{schwartz2020green, bender2021dangers}. Studies also quantify electricity usage and emissions, highlighting the importance of hardware efficiency \cite{lacoste2019quantifying, patterson2021carbon}. Optimizing model architecture, training procedures, and hardware efficiency are key strategies to mitigate environmental impact. 

\subsection{Combating the Impact}



Getzner et al. \cite{getzner2023accuracy} proposed an energy estimation pipeline to predict model energy consumption without execution. However, GPU measurement was not covered and CPU-based estimation has limitations. Unlike prior studies, our research provides a broad analysis from VGG16 to the latest vision transformers, extending beyond image classification to tasks like action recognition and semantic segmentation. Introducing a novel metric considering both model power consumption and traditional performance metrics aims to level the playing field, acknowledging hardware limitations faced by smaller organizations.

\subsection{Tracking Power Consumption}

Several open-source tools facilitate tracking emissions and electricity consumption. CodeCarbon \cite{codecarbon} is a Python package estimating CO2 emissions and electricity usage during code execution, including ML tasks. It integrates with PyTorch and TensorFlow, offering seamless quantification of emissions. CarbonTracker \cite{anthony2020carbontracker} monitors energy consumption throughout ML experiments, pinpointing carbon-intensive phases. Other tools like TraCarbon \cite{tracarbon} and Eco2AI \cite{eco2ai} are also popular. For our experiments, we utilize CodeCarbon, demonstrating a linear relationship between data and epochs/iterations (see Sec.~\ref{sec:data} for details). Authors can train on 1\% of data and 1 epoch to scale up estimated power consumption, simplifying the process.

\section{Proposed Metric}
\label{sec:metric}

Achieving high accuracy typically demands substantial computational resources, highlighting a trade-off between performance and environmental sustainability. To address this, there is a pressing need for a comprehensive metric that assesses both error rates and power consumption, promoting responsible resource use, energy efficiency, and fair model comparisons.

We introduce the Sustainable-Accuracy Metric (SAM)', detailed in Eq.~\ref{eq:gqi}. SAM considers both the accuracy rate (ranging from 0 to 1) and the total electricity used in kWh. Constants $\alpha$' and `$\beta$', set to 5 in our experiments, scale SAM to enhance understanding of its implications, as discussed in the Section~\ref{sec:alphabeta}. SAM calculates the cost of power per accuracy percentage, initially through a simple ratio of accuracy to power. To address the variability in power consumption across models, we use the logarithm of the electricity usage for better comparability. Furthermore, recognizing that not all accuracy gains are equivalent—enhancing performance from 80\% to 90\% is harder than from 30\% to 40\%—we adjust for this by exponentiating the accuracy value, acknowledging the increased challenge in achieving higher accuracy rates. This nuanced approach, acknowledging the Pareto principle \footnote{\url{https://en.wikipedia.org/wiki/Pareto_principle}}, ensures that SAM rewards methods that achieve exceptional accuracy.

\begin{equation}
    SAM = \beta \times \frac{accuracy^{\alpha}}{log_{10}(electricity)}
    \label{eq:gqi}
\end{equation}


Overall, SAM successfully penalizes high electricity consumption while emphasizing the value of accuracy improvements. This metric encourages research towards more sustainable models without compromising the contributions of models like vision transformers.

\subsection{Properties of the metric}

\noindent \textbf{Range}: Range for denominator is from [0, inf], range for numerator is from [0, inf], therefore the range for SAM is [0, inf].

\noindent \textbf{Penalizes High Power Consumption}: It penalizes models with high power consumption, promoting energy-efficient and environmentally responsible computing practices.
    
    \noindent \textbf{Promotes Accuracy}: While penalizing power consumption, the metric still encourages improvements in accuracy by considering the trade-off between error rate and energy use.
    
    \noindent \textbf{Non-Negative}: The metric is always non-negative, which is an important property for any evaluation metric, as negative values wouldn't make sense in this context.
   
    \noindent \textbf{Comparability}: It enables fair comparisons between different computational systems or models, facilitating informed choices based on efficiency and sustainability.
   
    \noindent \textbf{Sensitivity}: The metric is sensitive to variations in both error rate and power consumption, making it suitable for detecting small changes in system performance and power usage.

\section{Energy Consumption Across Tasks}
\label{sec:tasks}

We address a range of tasks encompassing image classification, semantic segmentation, and action recognition in videos. Given that the majority of these tasks employ an ImageNet-pretrained backbone, we designate ImageNet as the dataset for image classification. Additionally, we extend our analysis to encompass semantic segmentation and action recognition tasks, evaluating the overall costs by factoring in both pretraining costs and the expenses associated with training the models.

\section{Experimental Analysis}
\label{sec:exp}

\subsection{Image Classification}

Our study uses a wide range of architectural and cutting-edge models to evaluate performance comprehensively. These include established models like MobileNetv2 \cite{sandler2018mobilenetv2}, MobileNetv3 \cite{mobilenetv3}, ResNet \cite{resnet} series, VGG16 \cite{vgg16}, Inception v3 \cite{inceptionv3}, RegNetY \cite{regnet}, and DenseNet \cite{densenet}, as well as newer models such as EfficientNet v1 \cite{efficientnet}, v2 \cite{tan2021efficientnetv2}, and ViT \cite{vit}. We also explore novel approaches like Swin Transformers \cite{swin}, DeIT \cite{deit}, DaViT \cite{davit}, BEiT \cite{beit}, EfficientFormer \cite{efficientformer}, and ConvNeXt \cite{convnext}, with adaptations from pretraining databases like ImageNet21k \cite{im21k} and JFT \cite{jft}. Advanced techniques like CLIP \cite{clip} and MAE \cite{mae} are included to ensure a broad evaluation spectrum. We use the Pytorch Image Models (timm) \footnote{\url{https://huggingface.co/timm}} for all implementations.

\begin{figure*}
    \centering
    \includegraphics[width=0.99\textwidth]{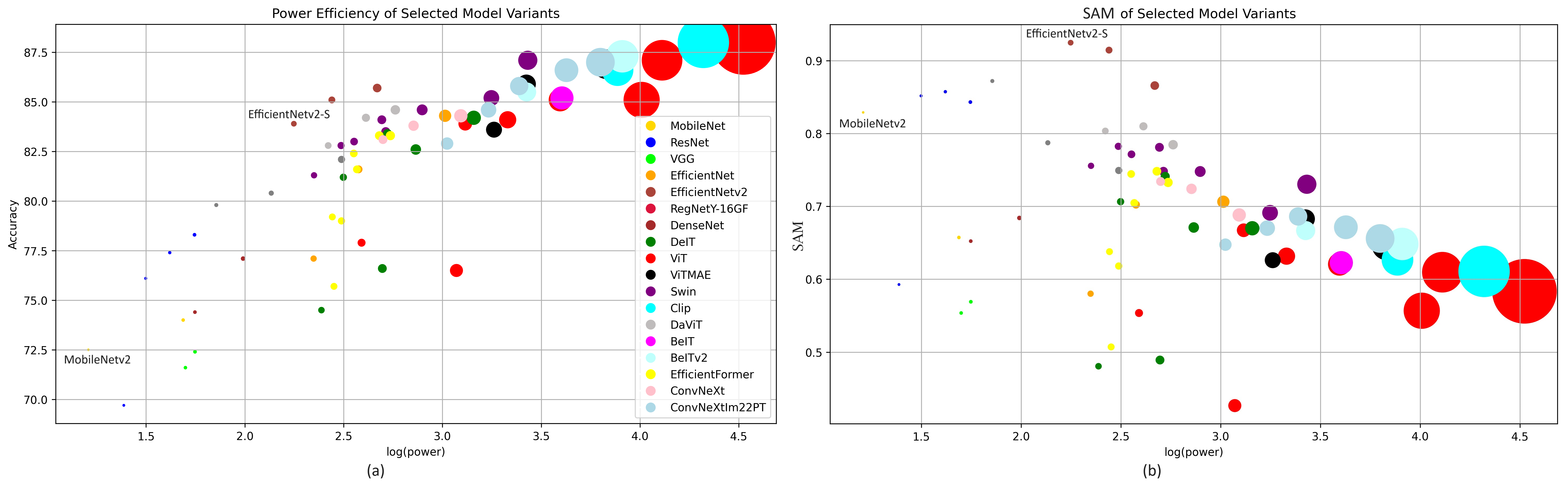}
\caption{Using SAM achieves a better balance between accuracy and electricity. In (a), lower accuracy models like MobileNet, ResNet, and EfficientNet are depicted poor compared to ViT-H. In (b), ViT-H and CLIP are penalized for high electricity usage, while EfficientNet and MobileNet rise. Marker points = electricity consumption (Zoom for details).}
    \label{fig:imgcls}
\end{figure*}

Figure \ref{fig:imgcls} illustrates the trade-off between model accuracy and power consumption, using 1 NVIDIA RTX A6000 48 GB GPU, with 10\% of the ImageNet \cite{imagenet} dataset, a fixed batch size of 32, and other consistent hyperparameters from the source papers. Figure \ref{fig:imgcls} (a) plots model accuracies against their power consumption logarithms, showing higher accuracy at higher power costs. Figure \ref{fig:imgcls} (b) introduces the SAM metric, which favors models like MobileNet and EfficientNet for their efficiency, crucial for sustainability and competing against well-resourced companies. 


\subsection{Semantic Segmentation}

Our research includes experiments on a variety of semantic segmentation methods. We explore classical approaches like PSPNet \cite{pspnet} and DeepLabv3 \cite{deeplabv3}, both using a ResNet101 backbone, as well as the real-time model BiseNet \cite{bisenet} with a ResNet18 backbone. ViT-based models Segmenter \cite{segmenter} and SETR \cite{SETR}, both with ViT-L backbones, and the universal model Mask2Former \cite{mask2former} with a ResNet50 backbone are tested alongside self-supervised models BEiT \cite{beit} and MAE \cite{mae} using ViT-B backbones.

Experiments are conducted over 2k iterations on 4 NVIDIA V100 32GB GPUs, with a batch size of 8, utilizing MMsegmentation \footnote{\url{https://github.com/open-mmlab/mmsegmentation}} and datasets Cityscapes \cite{cityscapes} and ADE20K \cite{ade}. FLOPS are calculated for an input size of $3 \times 512 \times 512$. Table \ref{tab:imgseg} presents a  efficiency analysis comparing these models. 2D CNN backbones are found to be more power-efficient than ViT backbones but maintain competitive or superior performance. Our SAM analysis highlights the efficiency advantages of these models.

\begin{table*}[]
\centering
\resizebox{.9\textwidth}{!}{
\begin{tabular}{cccccccc}
                                                &    & \multicolumn{3}{c}{Train}                                                                     & \multicolumn{3}{c}{Test}                                                                  \\
                                                    \hline
\multicolumn{1}{c}{Dataset} & \multicolumn{1}{c}{Models}                          & \multicolumn{1}{c}{GFLOPs} & \multicolumn{1}{c}{Parameters} & \multicolumn{1}{c}{Electricity} & \multicolumn{1}{c}{mIoU} & \multicolumn{1}{c}{SAM}      & \multicolumn{1}{c}{Electricity} \\ \hline
\multirow{6}{*}{Cityscapes} & PSPNet                                  & 256G& 65.6M& 67.9001 & 80.2                           & 0.906& 0.019259      \\         
& DeepLabv3                                  & 348G& 84.7M& 79.5025& 81.3& 0.935& 0.031512\\    
& BiSeNet                                  & 14.8G& 13.3M& 27.0179& 77.7& 0.989& 0.004970\\    
& Segmenter                                  & 400G& 334M& 1243.6126& 81.3& 0.574& 0.074713\\    
& SETR                                   & 417G& 310M& 1243.4491& 81.6& 0.585& 0.073880\\   
& Mask2Former                                   & 90G & 63M & 48.5331& 82.2& 1.113& 0.005852\\   
\hline
\multirow{2}{*}{ADE20K} & BEiT  & 605G& 162M& 1562.5031& 45.6& 0.031& 0.053533\\  
& MAE  & 605G& 162M& 1902.9770& 48.1& 0.039& 0.035615\\  
\hline
\end{tabular}}
\caption{A comparison of electricity consumed, accuracy and the proposed metric over multiple models. When we look at models using 2D CNN backbones, we see that they use much less electricity compared to ViT backbones. They can still compete with and sometimes even outperform ViT models. Our SAM analysis confirms this by ranking these methods much higher in terms of efficiency and performance.}
\label{tab:imgseg}
\end{table*}

\subsection{Action Recognition}

In this section, we evaluate the performance of models in the video action recognition task. Our experiments include various models: TimeSformer \cite{gberta_2021_ICML}, a ViT-based model, classical CNN-based models like MoViNet \cite{kondratyuk2021movinets}, I3D \cite{carreira2017quo}, TSM \cite{lin2019tsm}, TRN \cite{zhou2017temporalrelation}, and the hybrid model UniFormerV2 \cite{Li_2023_ICCV} that combines ViTs and CNNs.

Experiments are conducted on 2 NVIDIA RTX 3090 24GB GPUs, with a batch size of 8, using the settings from the source papers and official implementations. We evaluate these models on Kinetics-400 \cite{DBLP:journals/corr/KayCSZHVVGBNSZ17} and Something-Something V2 \cite{goyal2017something} datasets.

Table \ref{tab:act} shows a power efficiency comparison of these models in action recognition, integrating the SAM to assess the trade-off between accuracy and electricity consumption. For the SSv2 dataset, TSM stands out for its low electricity usage and reasonable accuracy, leading to its high ranking. However, on the Kinetics dataset, where accuracy gaps are wider, TSM ranks lower.

\begin{table*}[]
\centering
\resizebox{0.9\textwidth}{!}{
\begin{tabular}{ccccccccc}
\hline
Dataset    &         & Timesformer & Uni V2 (IN21K) & Uni V2 (CLIP) & MoviNet & I3D & TSM & TRN \\
               \hline
&GFlops             & 590G      & 3600G  & 3600G     & 2.71G     & 65G      & 65G   & 42.94G    \\
&Params             &  121.4M   & 163.0M  & 163.0M     & 3.1M      & 12.1M    & 24.3M     & 26.64M    \\
\hline
\multirow{4}{*}{SSv2} &Train Ele     & 1328.0169 & 1421.4425  & 7614.0746 & 303.3792  & 118.4572 & 107.3610     & 100.5043    \\
&Acc           & 59.5      & 67.5  & 69.5      & 61.3      & 49.6     & 63.4      
 & 47.65    \\
&Test Ele      & 0.2245  & 0.2899  & 0.2899  & 1.9653  & 1.9604  & 0.5178    & 0.2569    \\
&SAM                & 0.119     & 0.222 & 0.209       & 0.174     & 0.072    & 0.252     
 & 0.061    \\
\hline
\multirow{4}{*}{K400} & Train Ele     & 1337.1037 & 1466.8697 & 7632.2455  & 421.3966  & 136.6454 & 124.6118    & ---    \\
&Acc           & 78.0      & 83.4   & 84.4     & 65.8      & 73.8     & 74.1      & ---     \\
&Test Ele      & 0.1790  & 0.2312  & 0.2312  & 1.5673  & 1.5634  & 0.4128    & ---    \\
&SAM                & 0.462     & 0.637  & 0.552     & 0.235     & 0.513    & 0.533     & ---   \\
\hline 
\end{tabular}}
\caption{A comparison of electricity consumption, accuracy, and the proposed metric across multiple models in the action recognition task on the Something-something v2 dataset and the Kinetics-400 dataset. Timesformer corresponds to the divided space-time attention and Uni V2 (IM21k) uses a ViT-B pre-trained on IM21k and similary Uni V2 (CLIP) corresponds to the pre-trained CLIP model. `Train Ele' and `Test Ele' corresponds to the electricity consumed at train and test time respectively. Using SAM we achieve a better trade-off.}
\label{tab:act}
\end{table*}

\subsection{Why is self-supervised learning useful?}

Self-supervised learning marks a major advancement in AI research, enabling models to learn from extensive unlabeled data, thus minimizing reliance on costly human annotations and expediting progress across various fields. This method enhances model performance, transferability, and efficiency in diverse applications by generating versatile representations suitable for multiple downstream tasks. In our studies, we investigate two prominent self-supervised methods: BEiT and MAE. Despite their resource intensity during initial training, these methods are directly applicable to multiple tasks, such as image classification, semantic segmentation, and object detection (specifically reported by MAE).

\begin{figure*}
    \centering
    \includegraphics[width=0.75\textwidth]{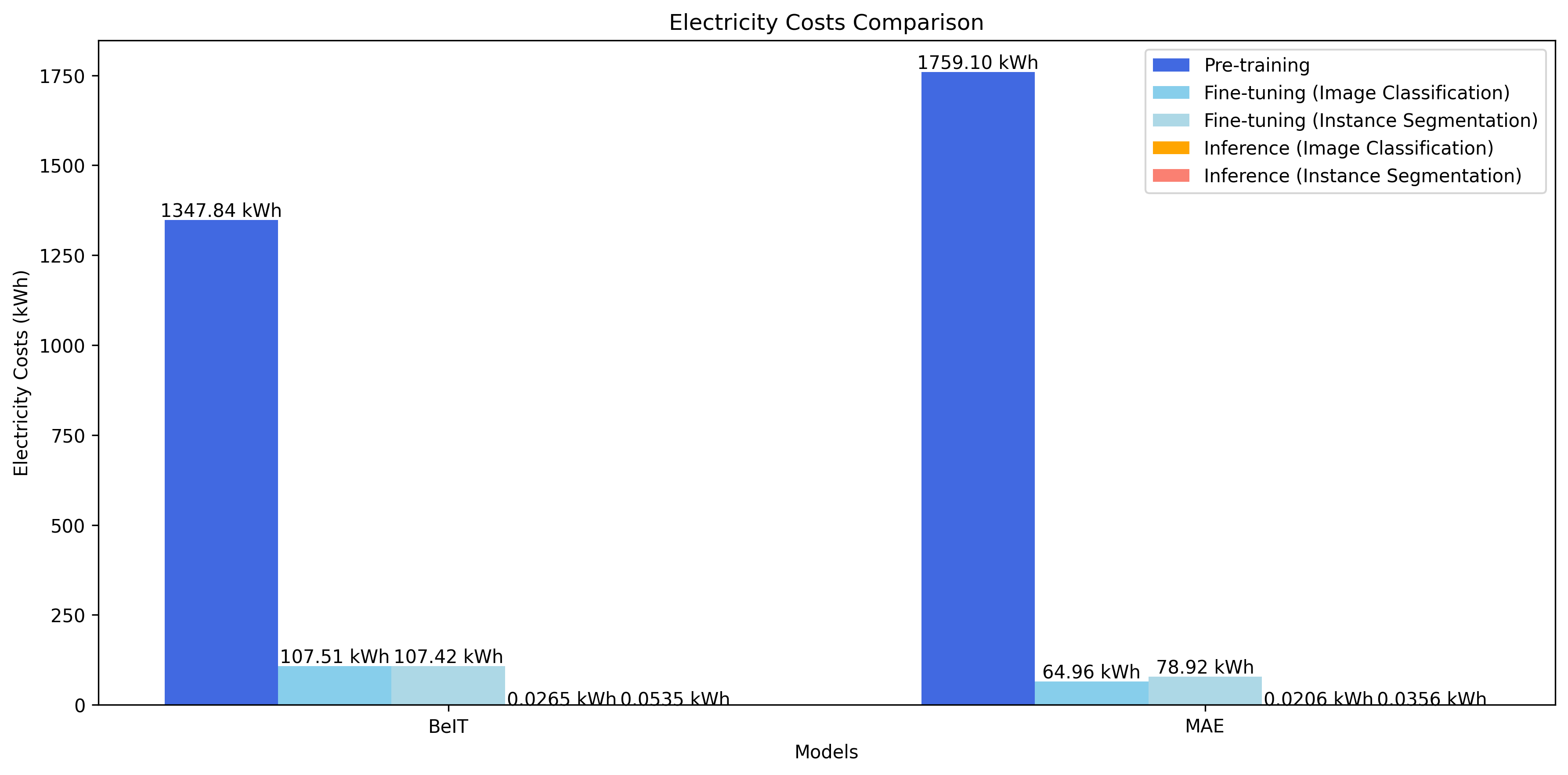}
    \caption{Comparing the cost of self-supervised pre-training and then fine-tuning along with inference of BEiT and MAE in terms of electricity consumption. The inference cost is so low in comparison that it is not even visible in the graph.}
    \label{fig:self}
\end{figure*}

Figure \ref{fig:self} illustrates that over 90\% of total training expenses are due to self-supervised pre-training. This highlights the cost-efficiency of fine-tuning self-supervised methods for various tasks or datasets, emphasizing their substantial benefits in real-world scenarios. The robust and transferable features produced by self-supervised learning are extremely valuable, yet the high training costs pose accessibility challenges for smaller entities, raising concerns about competitive disparities.

\subsection{Data vs Power Consumption}

\begin{figure*}
    \centering
    \includegraphics[width=0.75\textwidth]{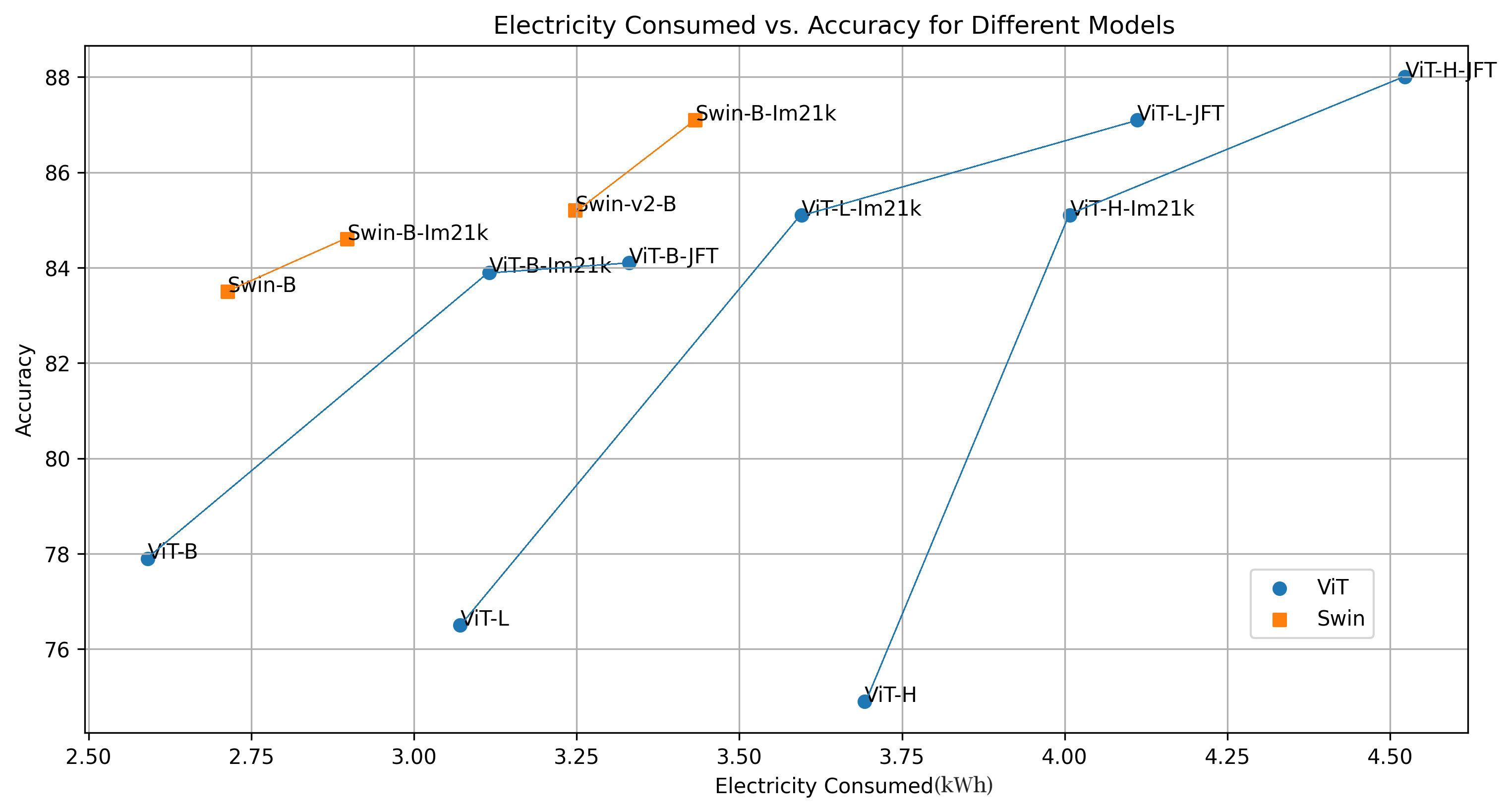}
    \caption{Comparing the effect of pre-training on Swin and ViT in terms of accuracy and electricity consumption.}
    \label{fig:pre-train}
\end{figure*}

Pre-training on extensive datasets notably leads to significant electricity consumption. This raises concerns, especially given the limited public availability of datasets like JFT or those used by CLIP, and the lack of accessible hardware for such intensive pre-training tasks. Moreover, the environmental impact of these pre-training processes prompts a critical evaluation: is large-scale pre-training justifiable? To explore this, we analyze improvements in models such as Swin (trained on Im21k) and ViT (trained on JFT). Our research, reveals a linear correlation between data volume and the electricity consumed during pre-training. This relationship is evident in Figure \ref{fig:pre-train}, which highlights performance changes linked to pre-training.

For example, ViT-B shows a 6\% improvement in performance over training from scratch while requiring fewer training epochs. However, this 6\% gain results in ten times the electricity usage compared to starting from scratch. This trade-off becomes more critical with larger ViT models, which often perform poorly when trained from scratch. While large-scale pre-training provides clear benefits in various tasks, with a primary focus on image classification in our study, the essential question is whether the electricity used is justified by the performance gains. Additionally, we question the fairness of comparing these improvements using datasets that are not publicly available.

\subsection{How much does hardware affect energy consumption?}

\begin{figure}
    \centering
    \includegraphics[width=0.8\linewidth]{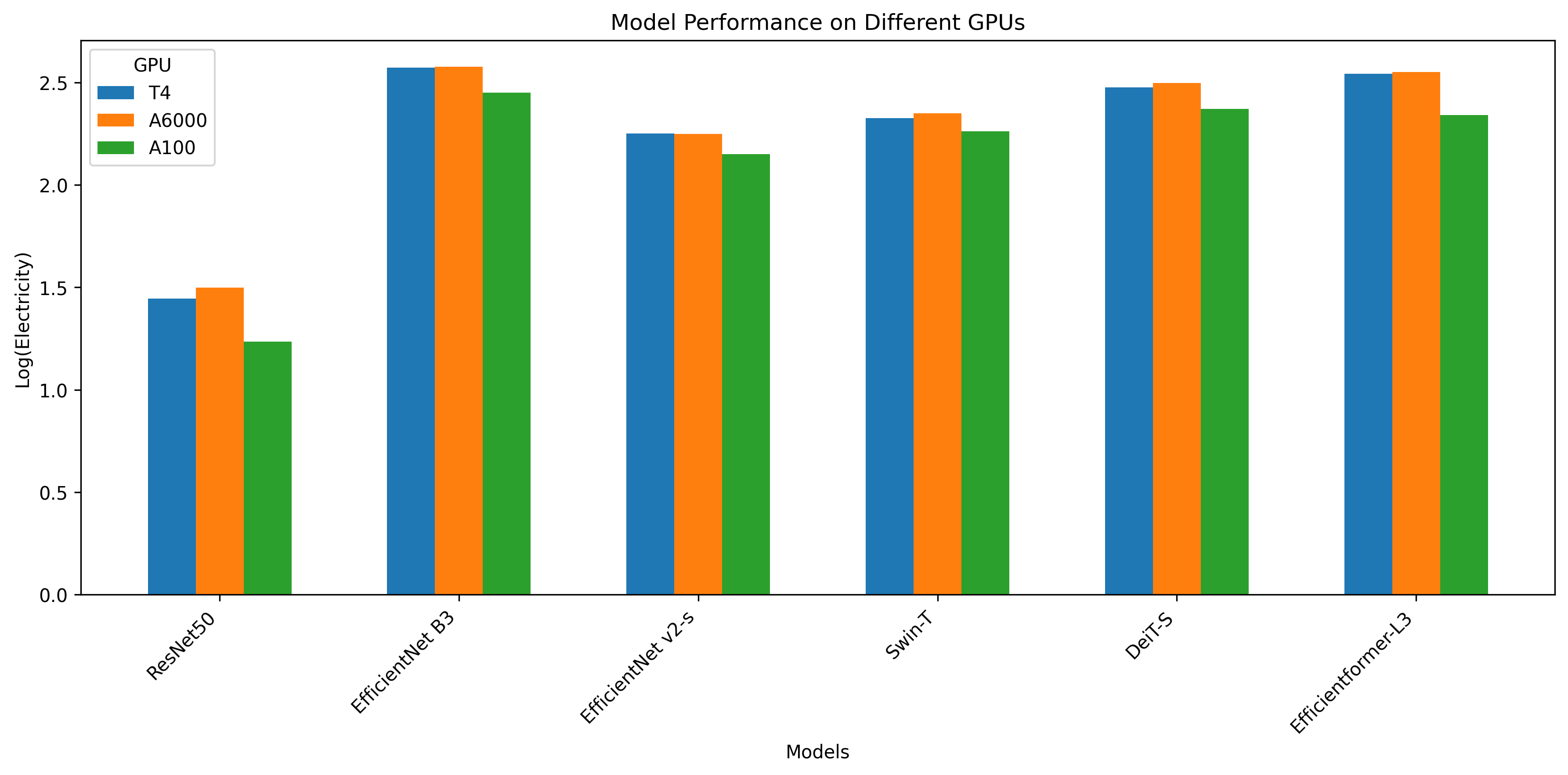}
    \caption{Model Performance Across GPUs: Despite Varied Hardware, Logarithmic Scaling Reveals Consistency}
    \label{fig:gpus}
\end{figure}

In evaluating the performance of various deep learning models on GPUs with different memory capacities, including the A100 with 40GB, Tesla T4 with 16GB, and A6000 with 48GB, an intriguing observation emerges. Despite the substantial variance in GPU memory, the utilization of a logarithmic scale, such as log(electricity) as a performance metric, effectively mitigates the discrepancies in raw performance metrics. This scaling allows for a fairer comparison, as it emphasizes relative improvements rather than absolute values. Consequently, even though the GPUs exhibit significant differences in memory size and hardware capabilities, the transformed data reveals that the models' performances remain remarkably consistent. This underscores the robustness of the chosen metric, which accommodates varying hardware configurations and ensures a reliable evaluation of deep learning model performance across diverse computing environments. This can be seen in Figure \ref{fig:gpus}.

\subsection{How do hyperparameters affect energy consumption?}

\begin{figure}
    \centering
    \includegraphics[width=0.8\linewidth]{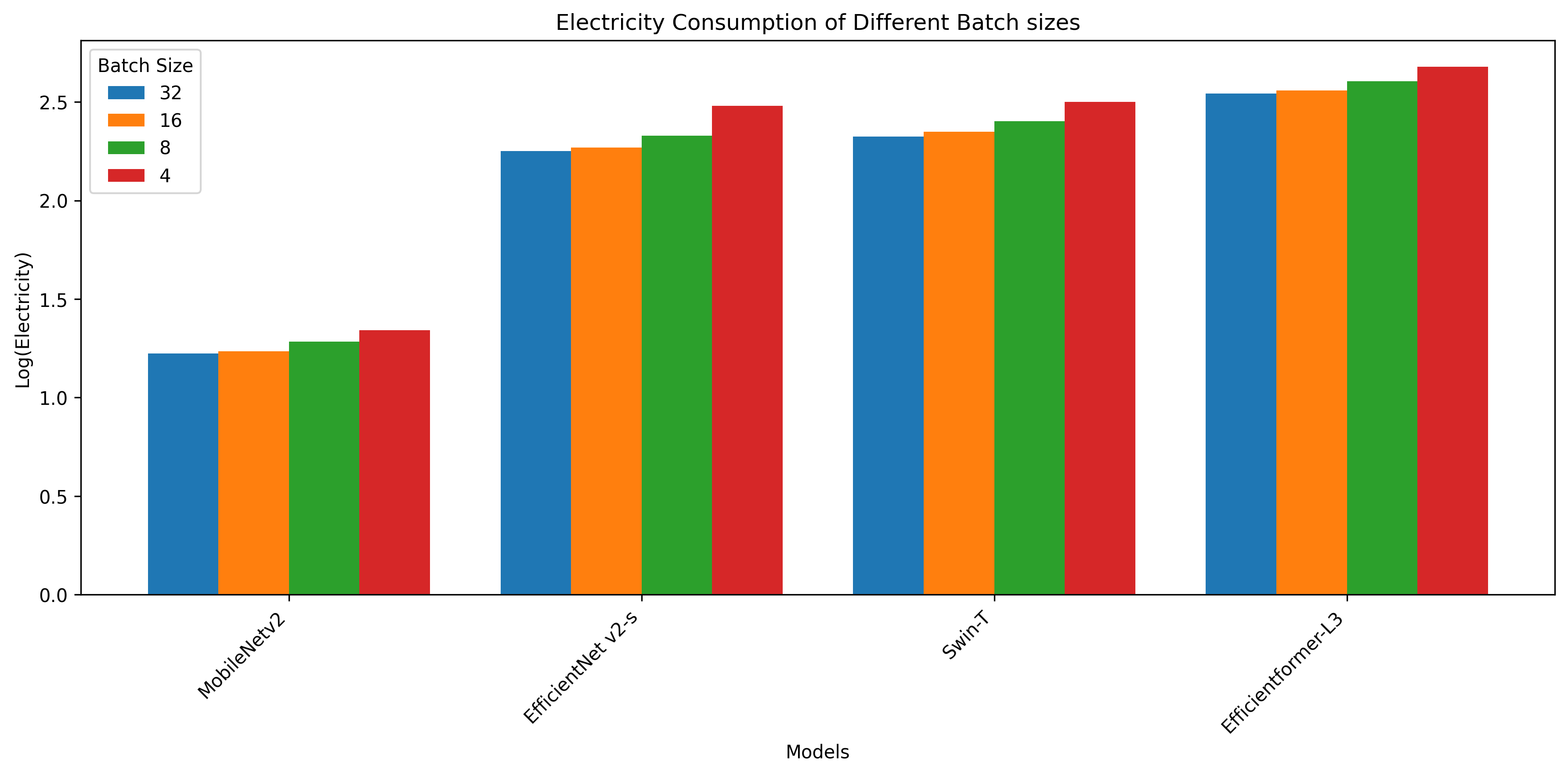}
    \caption{Model Performance Across Batch Sizes: The Need for Fixed Batch Sizes is Necessary for Fair Comparison.}
    \label{fig:batch}
\end{figure}

To assess the electricity consumption of popular deep learning models, namely MobileNet v2, EfficientNet v2-s, Swin-T, and Efficientformer-L3, we conducted a comprehensive analysis by varying the batch sizes during testing. Our findings highlight an expected trend: as batch sizes decrease, electricity consumption increases. This phenomenon is rooted in the intricate dynamics of deep learning processes. Smaller batch sizes lead to reduced parallelization, resulting in longer training times and, consequently, higher electricity consumption. Notably, when examining these differences in electricity consumption on a logarithmic scale, the variations appear relatively modest. Nevertheless, it is imperative to emphasize that, for fair model comparisons using our metric, a fixed batch size must be maintained. This standardization is essential because our metric's basis is tied to electricity consumption, and any alteration in batch size could skew the results. In summary, while batch size alterations do influence electricity consumption, our metric's integrity hinges on maintaining a consistent batch size for equitable model evaluations. This is shown in Figure \ref{fig:batch}.

\subsection{How to scale up for quicker reporting of results?}

One challenge associated with this metric is its susceptibility to hardware variations, where the efficiency may vary depending on the hardware used (for instance, TPUs might offer better performance, or larger GPUs might allow for larger batch sizes). Consequently, if we were to assess our models against this metric, we would need to re-run them multiple times using different hardware configurations. However, our research demonstrates that this dependency is actually linear with respect to the number of epochs and the percentage of data utilized. This means that we can initially run each model for just one or five epochs on a small subset of data, such as 1\% or 10\%, and then easily extrapolate the results to the entire dataset and the specified number of epochs mentioned in each research paper.

\subsection{Dependency of energy consumption with number of epochs}

\begin{figure}
    \centering
    \includegraphics[width=0.6\linewidth]{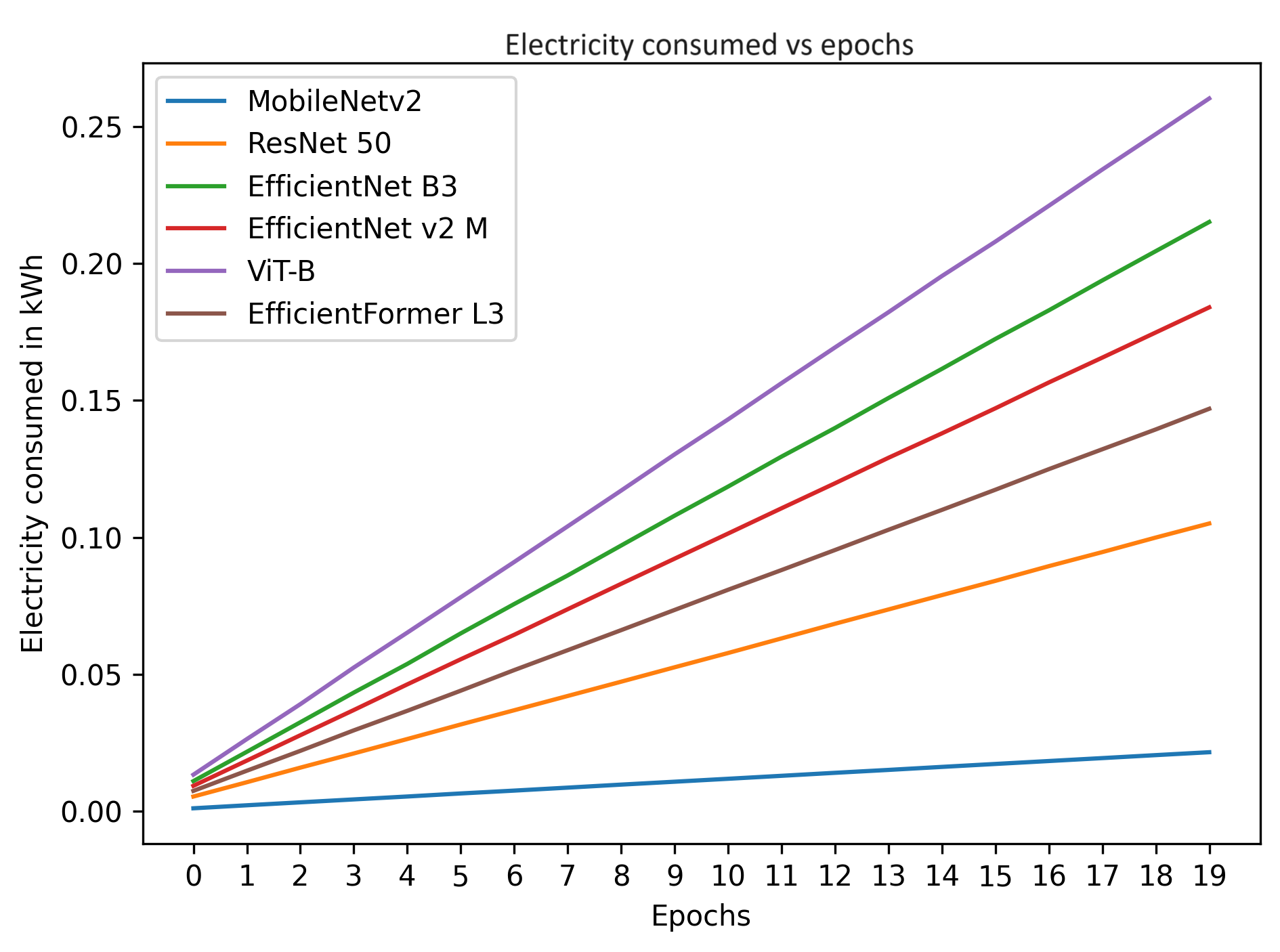}
    \caption{Electricity Consumed Across Epochs: Linear Relationship Observed.}
    \label{fig:epoch}
\end{figure}

We run MobileNet v2, ResNet50, EfficientNet B3, EfficientNetv2 M, ViT-B and EfficientFormer-L3 for a total of 20 epochs and calculate electricity consumption at each epoch. We plot this in Figure \ref{fig:epoch}. We see an approximately linear relationship and hence can easily scale up to any number of epochs. Further, since we use log scale in the metric, any small variations are further diminished. We use 1\% of the overall training data of ImageNet1k.

\subsection{Dependency of energy consumption with percentage of data}
\label{sec:data}

\begin{figure}
    \centering
    \includegraphics[width=0.6\linewidth]{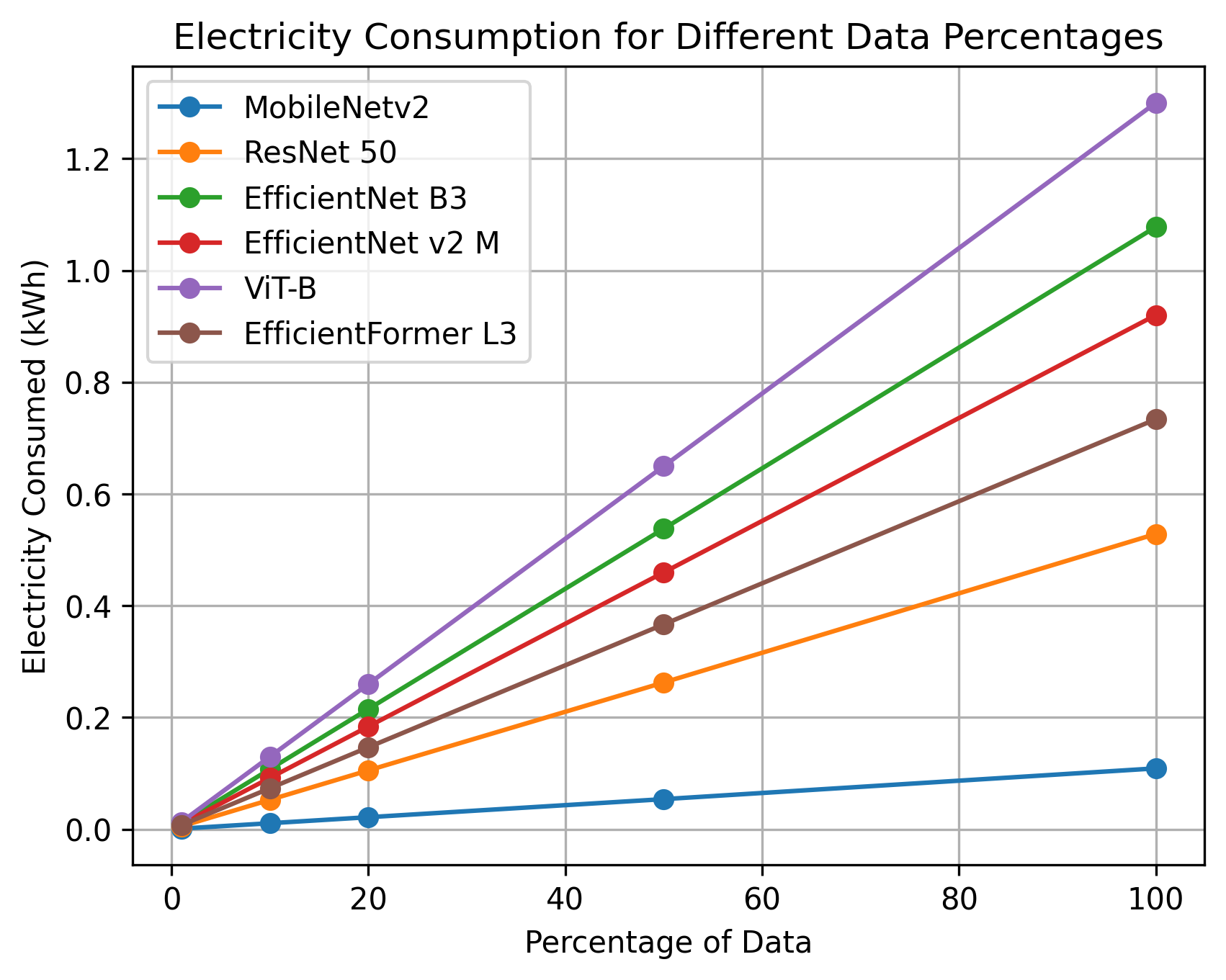}
    \caption{Electricity Consumed Across Different Percentages of Data: Linear Relationship Observed.}
    \label{fig:data}
\end{figure}

We run MobileNet v2, ResNet50, EfficientNet B3, EfficientNetv2 M, ViT-B and EfficientFormer-L3 for a total of 20 epochs and calculate electricity consumption at each epoch, but in this case we run them on varying amounts of data starting from 1\% then 10\%, 20\%, 50\% and finally 100\% of the data. We plot this in Figure \ref{fig:data}. We see an approximately linear relationship and hence can easily scale up to 100\% of the data with just 1\% of the data. Further, since we use log scale in the metric, any small variations are further diminished. 

\subsection{Why approximations work?}

The practice of approximating electricity consumption values by running models on a fraction of the dataset and for a fraction of the total epochs is efficient and viable due to several key factors. Firstly, the observed approximately linear relationships between energy consumption and both the number of epochs and the percentage of data used allow for straightforward extrapolation, enabling reliable estimates. Additionally, the use of a logarithmic scale in the energy consumption metric mitigates the impact of minor variations, ensuring the robustness of these approximations. Moreover, this approach significantly reduces computational load and time requirements, making it resource-efficient, especially when dealing with extensive datasets or rapid model configuration assessments. Lastly, the scalability of these approximations facilitates their application in scenarios involving larger datasets or extended training periods.

\subsection{Hyperparameters of the Metric}
\label{sec:alphabeta}

In the context of image classification, we established fixed values of $\alpha$ and $\beta$ as 5. What we observed was that, before setting $\alpha$ to 5, MobileNet v2 consistently achieved the highest SAM value, regardless of more recent research findings. This was primarily attributed to MobileNet v2's notably low power consumption. However, upon setting $\alpha$ to 5, we found that EfficientNet v2 surpassed all others in terms of SAM, while methods such as Swin and DeiT were competitive at a similar level. When it came to setting $\beta$, our focus shifted to examining the SAM values when $\beta$ was set to 1. Within this range, SAM values fluctuated between 0.111 and 0.185, presenting challenges in terms of interpretability. Consequently, we decided to increase the value of $\beta$ to 5. This adjustment led to a narrower SAM range, now spanning from 0.555 to 0.925, significantly enhancing the interpretability of the results. We ultimately opted to maintain these $\alpha$ and $\beta$ values consistently for both action recognition and semantic segmentation tasks, and our reported results in the tables reflect this choice.

\subsection{Limitations and Benefits}

Our metric calculation is based on the electricity required to train a model, which depends significantly on the hardware used. While we use a logarithmic scale to mitigate hardware impact, advancements in GPU technology could affect metric stability. Nevertheless, we demonstrate that extrapolating electricity costs is feasible due to a linear relationship between data volume, epochs, and power usage, allowing for quick assessments regardless of hardware variations. Another potential limitation is the interpretative use of hyperparameters $\alpha$ and $\beta$, which lack intrinsic meaning. Additionally, using a logarithmic scale for electricity gives greater weight to accuracy improvements, a choice made to avoid diminishing accuracy gains. A key strength of our approach is the ease of calculating electricity consumption across different platforms or servers, unlike more complex methods such as using Power Usage Effectiveness (PUE), which would require data center access. Despite its imperfections, we believe our metric promotes the development of more efficient models and represents a constructive step towards addressing the pressing issue of sustainable computing.

\section{Conclusion}

In conclusion, our study delves comprehensively into the critical intersection of deep learning model performance and energy consumption. While deep learning has unquestionably transformed numerous fields with its unparalleled accuracy, the escalating energy requirements of these models have given rise to environmental concerns and presented challenges for smaller research entities. To address this issue of escalating electricity consumption by models, we propose an innovative metric that places significant emphasis on accuracy per unit of electricity consumed. This metric not only levels the competitive field but also empowers smaller university labs and companies to compete effectively against their larger counterparts. Through an extensive examination of various deep learning models across different tasks, we have uncovered invaluable insights into the trade-offs between accuracy and efficiency. Our findings highlight the potential for more sustainable deep learning practices, where smaller, energy-efficient models can expedite research efforts while minimizing environmental impact. This research not only fosters a fairer research environment but also advocates for the adoption of efficient deep learning practices to reduce electricity consumption, ensuring a greener future for generations to come.
\bibliographystyle{splncs04}
\bibliography{eccvw}
\end{document}